\documentclass{article}
\usepackage{ijcai17}

\usepackage{times}
\usepackage{graphicx}
\usepackage{subfigure}
\usepackage{multirow}
\usepackage{amsmath}
\usepackage{amssymb}
\newcommand\numberthis{\addtocounter{equation}{1}\tag{\theequation}}

\title{Predicting Human Interaction via Relative Attention Model}
\author{Yichao Yan, Bingbing Ni, Xiaokang Yang\\
Shanghai Jiao Tong University, Shanghai, China  \\
\{yanyichao, nibingbing, xkyang\}@sjtu.edu.cn}

\begin{document}

\maketitle

\begin{abstract}
  Predicting human interaction is challenging as the on-going activity has to be inferred based on a partially observed video. Essentially, a good algorithm should effectively model the mutual influence between the two interacting subjects. Also, only a small region in the scene is discriminative for identifying the on-going interaction. In this work, we propose a relative attention model to explicitly address these difficulties. Built on a tri-coupled deep recurrent structure representing both interacting subjects and global interaction status, the proposed network collects spatio-temporal information from each subject, rectified with global interaction information, yielding effective interaction representation. Moreover, the proposed network also unifies an attention module to assign higher importance to the regions which are relevant to the on-going action. Extensive experiments have been conducted on two public datasets, and the results demonstrate that the proposed relative attention network successfully predicts informative regions between interacting subjects, which in turn yields superior human interaction prediction accuracy.
\end{abstract}

\section{Introduction}

Action prediction is defined as the problem of recognizing on-going activities based on temporally incomplete observations. It is a challenging task as only a part of the video is available for observation. Compared to individual action prediction, human interaction prediction is even harder, because the activities are more complex and involve more actors in the scene. More importantly, the incoming action of a subject might depend on the intention of the other subject, and this intention has to be inferred based on certain movement of this subject. In other words, to predict interaction, a good model should understand one subject's current action and how it will affect the other's response to this action in the near future.

Despite significant progress in the past few years, human interaction prediction is still challenging mainly due to the following two unanswered questions. The first one is how to model interaction or relative information. Second is how to discover the most discriminative regions and make use of them to make prediction. Solving these two difficulties will always bring performance gain over holistic or global feature learning methods.

\begin{figure}
\begin{center}
\includegraphics [width=\linewidth] {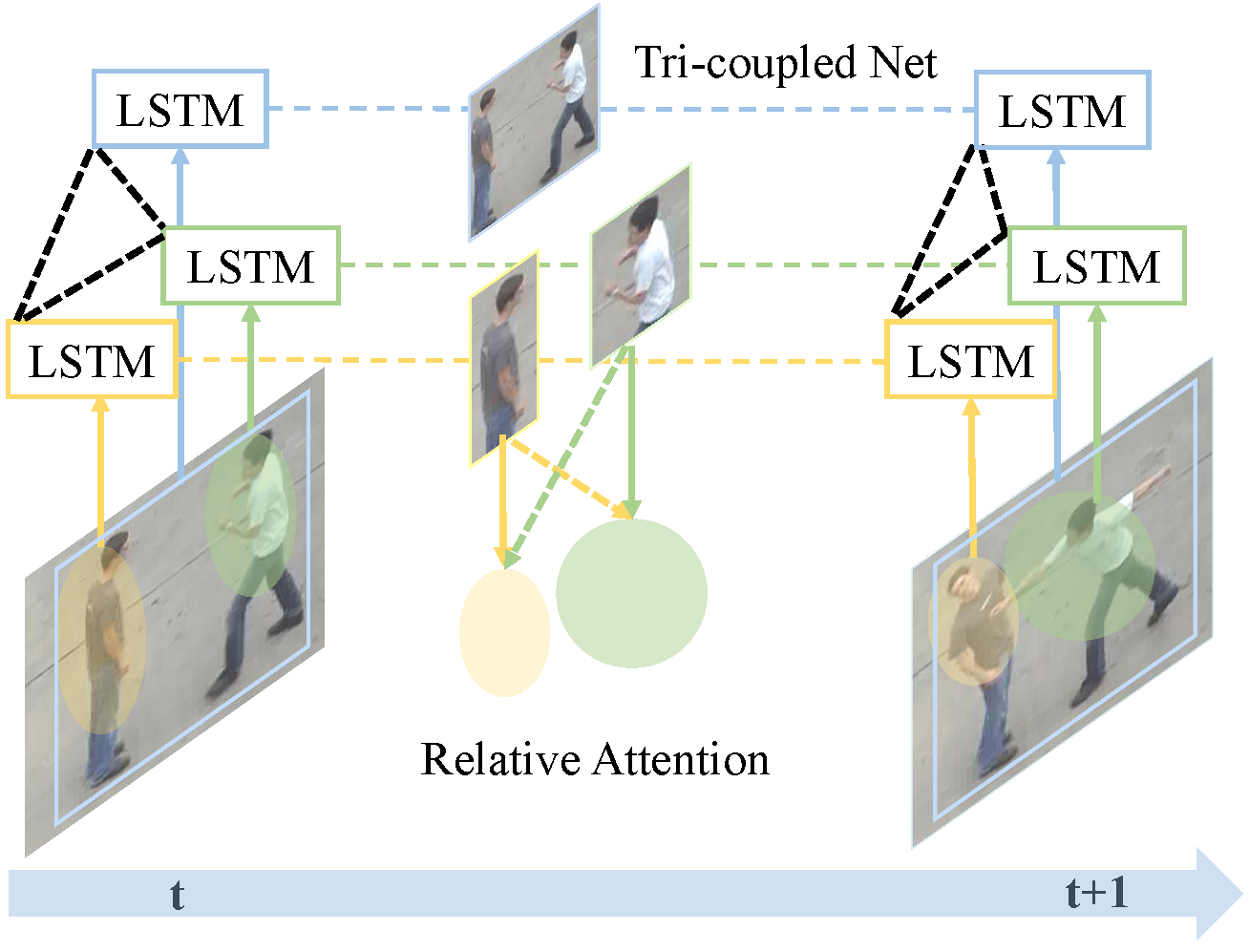}
\vspace{-9mm}
\caption{
Overview of our framework. We design a tri-coupled deep recurrent structure representing both interacting subjects and global interaction status, and embed an attention module to predict the discriminative regions of each subject.
}
\label{fig:overview}
\vspace{-7mm}
\end{center}
\end{figure}

However, previous methods do not address these questions in a proper way. Recent methods mainly resort to: (1) holistic representation \cite{DBLP:conf/cvpr/CaoBBNYMLDSW13,DBLP:conf/iccv/Ryoo11,DBLP:conf/eccv/KongKF14};
(2) individual representation \cite{DBLP:conf/eccv/LanCS14} and (3) discriminative part based representation \cite{DBLP:conf/iccv/XuQM15,DBLP:conf/mm/Chen15a}.
Despite their favorable performance on recent benchmark datasets \cite{UT-Interaction-Data,DBLP:conf/eccv/KongJF12}, we have the following observations on their limitations.
First, holistic feature based methods \cite{DBLP:conf/cvpr/CaoBBNYMLDSW13,DBLP:conf/iccv/Ryoo11,DBLP:conf/eccv/KongKF14} usually encode the whole scene into a global feature vector, the richer information contained in individual subjects is ignored.
Second, although individual representation based method \cite{DBLP:conf/eccv/LanCS14} models both interactive subjects, they are usually modeled separately. How to effectively model their relationship is not well exploded.
Third, discriminative part based methods \cite{DBLP:conf/iccv/XuQM15,DBLP:conf/mm/Chen15a} try to select discriminative patches/parts to represent the actions. Such discriminative patch/part detectors usually apply to the video frame-by-frame, thus the detected patches/parts are not temporally consistent. Moreover, such methods are hard to distinguish similar movements, as the generated patches are also similar.

To explicitly address the above issues, we propose a tri-coupled relative attention framework.
On one hand, a \textbf{tri-coupled interaction fusion network} is proposed to model mutual influence between subjects involved in the interaction. This network is composed of three recurrent sub-structures, which accept three streams of information representing both interacting subjects and the global interaction region enclosing both subjects. To capture the dependency between subjects, at each time-step, information flows from all three streams are aggregated to the hidden node of the current time-step, and then output the new status information for both interacting subjects. We make two remarks. First, we denote it by \emph{coupled recurrent network} because status information of one subject is linked to the other stream, in order to assist the prediction of the next status of the other subject. Second, information extracted from the global scene (which encloses both subjects) is also utilized to predict the interaction status of both subjects.
In this way, both local motion information and global motion information are fully utilized, which are complements to each other.
On the other hand, built on this tri-coupled recurrent infrastructure, we introduce a \textbf{relative attention network}. The motivation is that some local motion (attended small regions) might give very useful information to predict the other subject's response in the future. For example, if a person extends his arm or leg, another person is likely to dodge, a punching/kicking is more likely to happen. In this situation, the arm/leg region is crucial for predicting another person's response. Motivated by this observation, a visual attention module is embedded to the recurrent structure to predict the discriminative regions of each subject. At each time-step, the attention module receives information from both interactive subjects, as well as their hidden states of previous time-step, and then output the attended regions of both subjects. In this way, only the attended regions are input into the recurrent networks, providing discriminative local information.

The proposed network is extensively compared with some popular methods for encoding human interaction on two popular datasets, the results of the proposed method show favorable performance against the state-of-the-art methods.

\section{Related Work}
\textbf{Traditional methods.}
For action prediction, many previous works focus on finding good feature representation (usually bag-of-words features or sparse coding) and training SVM-like classifiers. For example, Ryoo~\shortcite{DBLP:conf/iccv/Ryoo11} proposes two BoW based representation, i.e., the integral bag-of-words (IBoW) and dynamic bag-of-words (DBoW).
Cao et al.~\shortcite{DBLP:conf/cvpr/CaoBBNYMLDSW13} apply sparse coding to derive the activity bases, and use the reconstruction error in the likelihood computation.
Lan et al.~\shortcite{DBLP:conf/eccv/LanCS14} propose a hierarchical representation and combine it with a max-margin learning framework for action prediction.
Another two max-margin frameworks \cite{DBLP:conf/eccv/KongKF14,DBLP:conf/cvpr/NguyenT12} are built upon structured SVM model, but extend it to accommodate sequential data.
Kong and Fu~\shortcite{DBLP:journals/pami/KongF16} further extend this framework using compositional kernels to model the relationship of partial observations.
Xu et al.~\shortcite{DBLP:conf/iccv/XuQM15} consider action prediction as a query auto-completion problem.
These methods use hand-crafted features and encoding methods to represent the video. The difference of our work lies in the using CNN/LSTM features rather than hand-crafted features, which enables our model to be trained end-to-end.

\textbf{CNN based methods.}
Many CNN based methods have been focused on activity recognition and video classification. In \cite{Ji20133D}, a 3D CNN model is proposed for action recognition.
Karpathy et al.~\shortcite{DBLP:conf/cvpr/KarpathyTSLSF14} explore several approaches for fusing information over temporal dimension trough the CNN, but only achieving marginal improvement than the single frame baseline, which indicates that learning motion information is difficult for CNN.
To address this issue, Simonyan and Zisserman~\shortcite{DBLP:conf/nips/SimonyanZ14} propose a two-stream architecture which directly incorporate motion information from optical flows, achieving significant improvements compared to previous CNN based methods.
However, such approaches are based on single frames, not able to represent long-term temporal clue.

\textbf{RNN based methods.}
Recurrent neural network (RNN) and Long Short Term Memory (LSTM) \cite{DBLP:journals/neco/HochreiterS97} are powerful tools to model sequential data.
LSTMs have been applied to action classification in \cite{DBLP:conf/icann/BaccoucheMWGB10,DBLP:conf/cvpr/DonahueHGRVDS15}.
The work of \cite{DBLP:conf/mm/WuWJYX15,DBLP:conf/cvpr/NgHVVMT15} further improves the performance by building a hybrid model incorporating both spatial and temporal clue.
Ibrahim et al.~\shortcite{Ibrahim_2016_CVPR} build a 2-stage deep temporal model for group activity recognition.
Ma et al.~\shortcite{Ma_2016_CVPR} design novel ranking losses for training LSTM which enforce the score margin between the correct and incorrect categories to be monotonically non-decreasing.
Visual attention model is also investigated for action recognition in \cite{DBLP:journals/corr/SharmaKS15}.
Song et al~\shortcite{DBLP:journals/corr/SongLXZL16} build a spatio-temporal attention model from skeleton data. These works mainly focus on recognizing action of a single object or group activity, they achieve promising result when the complete video is observed. In contrast, our framework is explicitly designed for person interaction involving a pair of persons in the scene, and it still achieves satisfactory results when only a small part of the video is observed.

\section{Methodology}
The problem is formulated as follows. We denote a complete video of duration $T$ as $V[1:T]$, the task is to predict the action $\rm{y}$ with only partial observation $V[1:t], t\in\{1,...,T\}$, the observation ratio is $\frac{t}{T}$. The complete videos are only accessible for training, and the performance is evaluated by calculating the prediction accuracy with a fixed observation ratio for all the test videos. In this work, we assume the bounding box of each person and the global scene enclosing the two actors are located in each frame.
In the rest of this section, we use $\mathbf{X}$ to denote the CNN feature extracted from raw frames. $\mathbf{L}$ denotes the attention weights corresponding to the attended region. The inputs and hidden states of LSTM network are denoted as $\mathbf{x}$ and $\mathbf{h}$ respectively. The weights and bias terms in our networks are denoted as $\mathbf{W}, \mathbf{U},\mathbf{V}$ and $\mathbf{b}$.

\begin{figure}
\centering
\subfigure[Global LSTM network.]{
\label{fig:subfig:a}
\includegraphics[width=3in]{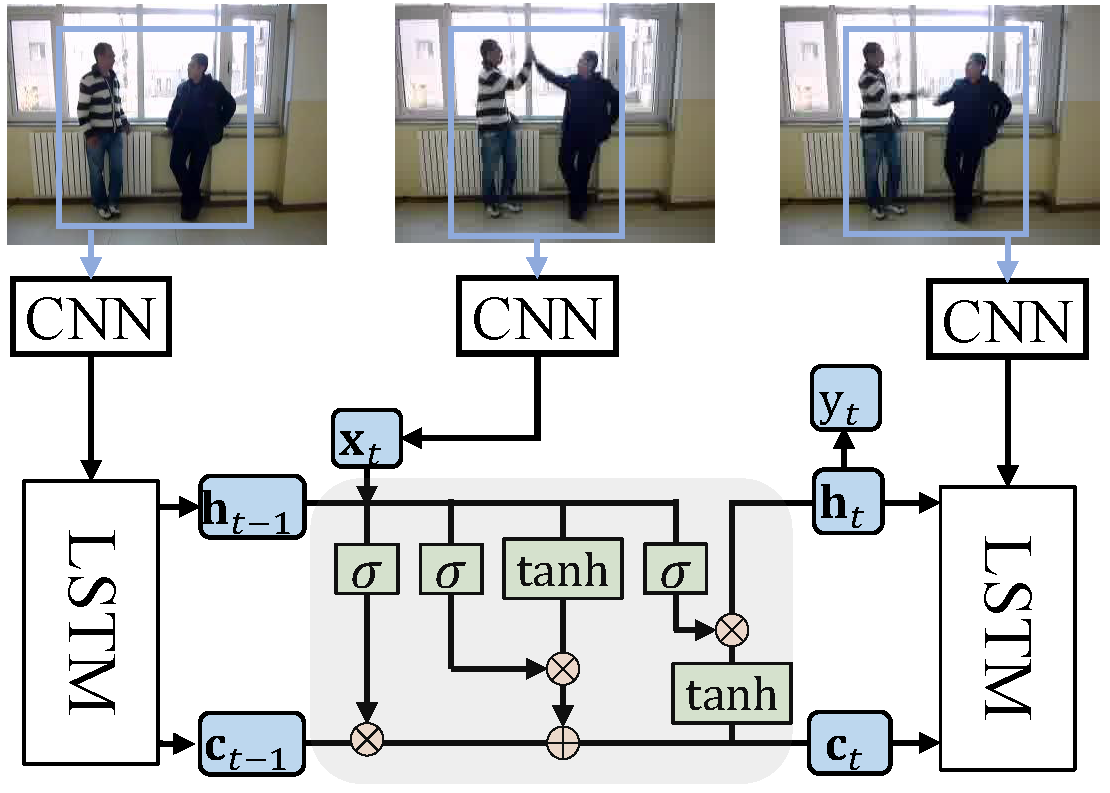}}
\subfigure[Naive fusion method.]{
\label{fig:subfig:b}
\includegraphics[width=3in]{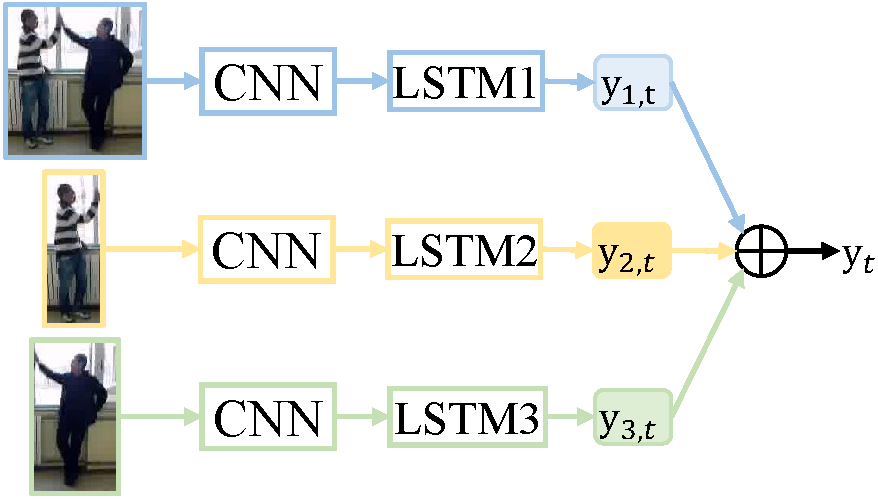}}
\vspace{-3mm}
\caption{
Two baseline methods for action prediction.
}
\label{fig:subfig}
\vspace{-3mm}
\end{figure}

\subsection{Tri-coupled Interaction Fusion Network}\label{sec:method}
With an LSTM network, information could be propagated from the first node to the last one, and the good nature of LSTM is very useful for our given task, i.e., to make full use of the observed information and make a prediction. Motivated by the success of recurrent neural networks in temporal sequence analysis, we employ LSTM network as our network prototype.
The frame-level features are input into LSTMs to model the spatio-temporal information.

In particular, each LSTM node includes three gates, (i.e., the input gate $\mathbf{i}$, the output gate $\mathbf{o}$ and the forget gate $\mathbf{f}$) as well as a memory cell.
At each time-step $t$, the input feature $\mathbf{x}_t$ and the previous hidden state $\mathbf{h}_{t-1}$ are input into the LSTM, as illustrated in Figure \ref{fig:subfig:a}. The LSTM network updates as follows:
\begin{align}
\hspace{-1mm}
{\mathbf{i}_t} &= \sigma({\mathbf{W}_i}{\mathbf{x}_t} +{\mathbf{U}_i}{\mathbf{h}_{t-1}}+{\mathbf{V}_i}{\mathbf{c}_{t-1}} + \mathbf{b}_i)&  \numberthis \label{eqn1}
 \\
{\mathbf{f}_t} &= \sigma({\mathbf{W}_f}{\mathbf{x}_t} + {\mathbf{U}_f}{\mathbf{h}_{t-1}}+{\mathbf{V}_f}{\mathbf{c}_{t-1}} + \mathbf{b}_f)&\\
{\mathbf{c}_t} &= \mathbf{f}_t\cdot{\mathbf{c}_{t-1}} + \mathbf{i}_t\cdot{\tanh({\mathbf{W}_c}{\mathbf{x}_t} + {\mathbf{U}_c}{\mathbf{h}_{t-1}}+ \mathbf{b}_c)}&\\
{\mathbf{o}_t} &= \sigma({\mathbf{W}_o}{\mathbf{x}_t} + {\mathbf{U}_o}{\mathbf{h}_{t-1}}+{\mathbf{V}_o}{\mathbf{c}_{t}} + \mathbf{b}_o)&\\
{\mathbf{h}_t} &= \mathbf{o}_t\cdot{\tanh({\mathbf{c}_t})} \numberthis \label{eqn5}
\end{align}
where $\sigma$ is the sigmoid function and $\cdot$ denotes the element-wise multiplication operator. $\mathbf{W}_{*}$, $\mathbf{U}_{*}$ and $\mathbf{V}_{*}$ are the weight matrices, and $\mathbf{b}_{*}$ are the bias vectors. The memory cell $\mathbf{c}_t$ is a weighted sum of the previous memory cell $\mathbf{c}_{t-1}$ and a function of the current input. The weights are the activations of forget gate and input gate respectively.

For the task of interaction prediction, the most straightforward idea is to model the global interaction regions enclosing both subjects with a single LSTM network, as other activity recognition system \cite{DBLP:conf/cvpr/DonahueHGRVDS15}. We denote it a \textbf{global LSTM network}, which takes the complete region of action as input and models the global information of the observed video. As shown in Figure \ref{fig:subfig:a}, the frame-level features are extracted by a CNN extractor, and then input into the LSTM network for classification. Here, we use Alexnet \cite{DBLP:conf/nips/KrizhevskySH12} as CNN feature extractor.
The good nature of this structure is that all the information is modeled by a global LSTM, which is simple and effective for action recognition. However, the interaction of individual subjects is not explicitly modeled in the structure, the performance might be limited for the task of interaction prediction.

There are multiple options to model the mutual interactions of the interactive subjects. A naive approach is to model each subject with an individual LSTM model and then combine their predictions, which can be further enhanced by employing the prediction of the global LSTM network. We denote this structure as \textbf{naive fusion network}, see Figure \ref{fig:subfig:b}.
This structure employs both global and local interactive information, but it also suffers from a major limitation. Some subjects are likely to have very similar behaviours in different interactions, e.g., the \textit{dodge} action in both \textit{kick} and \textit{box}. The prediction scores of these subjects can be very confusing, directly summing up their prediction scores may bring side effects to the overall results.

To address this issue, we design a joint training scheme that simultaneously models the interactive state of the two subjects. In particular, each subject is also represented by an LSTM network, but the hidden states of the two LSTMs are shared at each time-step.
In this case, the terms $\mathbf{U}_{*}{\mathbf{h}_{t-1}}$ in Equation \ref{eqn1} to Equation \ref{eqn5} are further represented by:
\begin{equation}
\mathbf{U}_{*}{\mathbf{h}_{t-1}}= \mathbf{U}_{*,s_1}{\mathbf{h}_{t-1,s_1}} + \mathbf{U}_{*,s_2}{\mathbf{h}_{t-1,s_2}},
\end{equation}
where $\mathbf{h}_{t-1,s_1}$is the previous hidden state of the network and $\mathbf{h}_{t-1,s_2}$ is the previous hidden state of the other subject.
This enables the information communication between the subjects, i.e., the statues of one subject can be used to help predict the action of the other subject.
Moreover, the outputs of the LSTMs are concatenated as a union feature for prediction, which is in contrast of combing the prediction scores of individual subject level LSTMs.
This structure allows the two the LSTMs to be trained together, i.e., there is a single loss for the networks.
We denote it a \textbf{coupled network}, see the top part of Figure \ref{fig:relative_attention}.

Although the coupled network explicitly models the spatio-temporal correlations of the two subjects, the global interactive information is not used in the structure. To integrate the global information into the network, we design a \textbf{Tri-coupled interaction fusion network}, as shown in the middle part of Figure \ref{fig:relative_attention}.
For the tri-couple structure, the LSTM representing the global interaction status is pre-trained as a vanilla LSTM, the other two LSTMs modeling the mutual interactions are modeled as:
\begin{equation}
\mathbf{U}_{*}{\mathbf{h}_{t-1}}= \mathbf{U}_{*,s_1}{\mathbf{h}_{t-1,s_1}} + \mathbf{U}_{*,s_2}{\mathbf{h}_{t-1,s_2}}+\mathbf{U}_{*,g}{\mathbf{h}_{t,g}},
\end{equation}
where $\mathbf{h}_{t,g}$ is the hidden state of global LSTM.

\subsection{Relative Attention Network}
\begin{figure}
\begin{center}
\includegraphics [width=\linewidth] {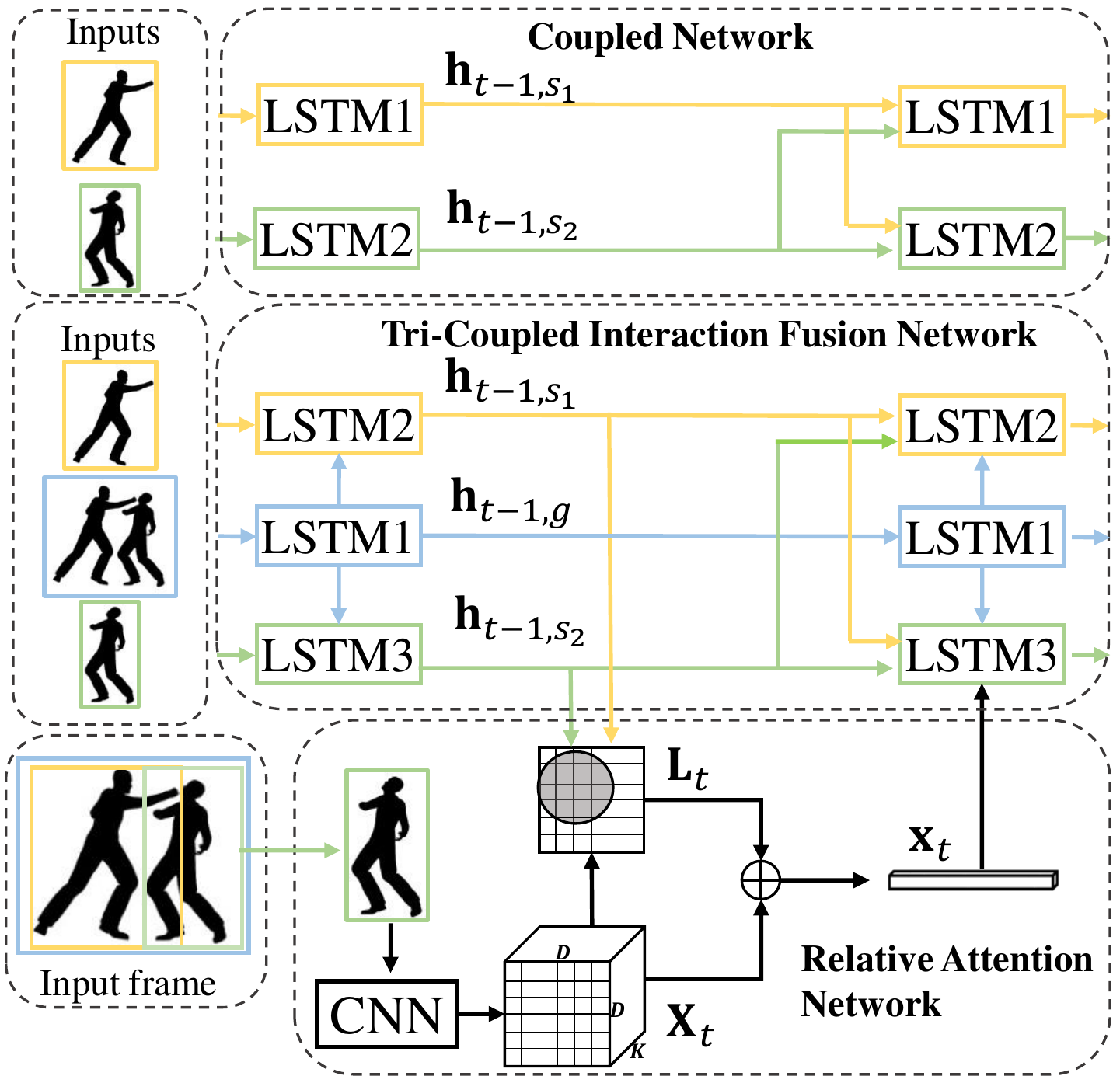}
\vspace{-8mm}
\caption{
Illustration of the Coupled Network, Tri-coupled Recurrent Network and the Relative Attention Network. 
}
\label{fig:relative_attention}
\vspace{-5mm}
\end{center}
\end{figure}
For the task of action prediction, usually only a certain region is crucial for identifying an action.
Therefore, we would like our model to focus on these regions and to model the fine-grained details. Here, we embed our tri-coupled network with a relative attention module.

Two kinds of attention model have been used to address this issue. Hard attention \cite{DBLP:conf/nips/MnihHGK14,DBLP:journals/corr/BaMK14} samples attention location at each time stamp, which causes the system not differentiable. In contrast, soft attention \cite{DBLP:journals/corr/BahdanauCB14,DBLP:journals/corr/SharmaKS15} aims to learn a set of weights corresponding to each region, the model is differentiable and can be trained end-to-end using standard back-propagation.
Therefore, we adopt the soft attention model in our work.
Instead of extracting feature from the last fully connected layer, the soft attention model employs the last convolutional layer, resulting to $K$ convolutional maps of size $D*D$, which can be denoted as:
\begin{equation}
\mathbf{X}_{t}=\{\mathbf{X}_{t,1}, ...,\mathbf{X}_{t,D^2}\}, \qquad \textbf{X}_{t,i}\in\mathbb{R}^{K}.
\end{equation}
Specially, each vector $\mathbf{X}_{t,i}$ corresponds to a specific receptive field in the original image.

At each time-step $t$, we would like to assign weights to each location in the $D*D$ feature map. The attended region should have higher weights compared to less important regions. As each location in the feature maps corresponds to a certain receptive field in the original image, attending to the feature map plays the same role as attending to the original image. The attention weights $\mathbf{L}_{t}=\{l_{t,1},...,l_{t,D^2}\}$ at time-step $t$ is usually calculated using the following two features: the hidden state of the previous time-step $\mathbf{h}_{t-1}$ and the CNN feature map of the current time-step $\mathbf{X}_{t}$. See Figure \ref{fig:relative_attention}. The weights are normalized after a softmax layer:
\begin{equation}\label{eqa:9}
l_{t,i} = \frac{\exp{(\mathbf{W}_{h,i}{\mathbf{h}_{t-1}}+ \mathbf{W}_{X,i}{\mathbf{X}_{t}})}}{\sum_{j=1}^{D*D}\exp{(\mathbf{W}_{h,j}{\mathbf{h}_{t-1}}+ \mathbf{W}_{X,j}{\mathbf{X}_{t}})}} ,\\
\end{equation}
where $i\in{1,...,D^2}$ and $\mathbf{W}_{*,i}$ are the weights for the inputs. $l_{t,i}$ can be viewed as the probability of the $i$-th region to be important.
For the tri-coupled network, we can also take advantage of the mutual information to help locate the interesting region, i.e., to use the hidden states of neighboring LSTMs. The $\mathbf{W}_{h,i}{\mathbf{h}_{t-1}}$ term in Equation \ref{eqa:9} can be further decomposed into hidden state information from both subjects:
\begin{equation}
\mathbf{W}_{h,i}{\mathbf{h}_{t-1}} = \mathbf{W}_{h,i,s_1}{\mathbf{h}_{t-1,s_1}}+ \mathbf{W}_{h,i,s_2}{\mathbf{h}_{t-1,s_2}}.
\end{equation}
The final inputs for LSTM is a weighted summation of the attention vector $\mathbf{L}_t$ and the CNN features $\mathbf{X}_t$:
\begin{equation*}
\mathbf{x}_{t}=\sum_{i=1}^{D^2}l_{t,i}\mathbf{X}_{t,i}.
\end{equation*}

\subsection{Training the Network }
The proposed tri-coupled network and relative attention network can be jointly trained as a classification problem of $N$ classes ($N$ is the number of human interaction category). At each time-step, the hidden state $\mathbf{h}_t$ of each sub-LSTM is concatenated as the feature representation vector, which is further connected to a softmax layer. The output of the $N$-way softmax is the prediction of the probability distribution over $N$ different actions:
\begin{equation}
{\rm{y}_i} = \frac{\exp(\rm{y}'_i)}{\sum_{k=1}^{N}{\exp({\rm{y}'_k})}},
\end{equation}
where $\rm{y}'_j=\mathbf{w}_{j}\cdot\mathbf{h}_{t} +{b}_j$ linearly combines the LSTM outputs, and $\mathbf{w}$ and $b$ are the weight matrix and bias term of the softmax layer. The network is learned by minimizing $-\log{\rm{y}_k}$, where $k$ is the index of the true label for a given input. Stochastic gradient descent is used with gradients calculated by back-propagation.

\section{Experiments}
In this section, we present extensive experimental evaluations and in-depth analysis of the proposed method on the following two human interaction prediction benchmarks:

\textbf{UT dataset.} The UT-Interaction dataset (UTI) \cite{UT-Interaction-Data} contains videos of 6 classes of human-human interactions: shake-hands, point, hug, push, kick and punch. Except that point is a single action, all other activities are performed by a pair of actors. This dataset contains two subsets: UTI \#1 and UTI \#2. The backgrounds of UTI \#1 are mostly static with little camera jitter, while the backgrounds of UTI \#2 are moving slightly and containing more camera jitters. Both of the two subsets contain 10 videos of each interaction class. We adopt 10-folder leave-one-out cross validation setting to measure the performance of the two subsets.

\textbf{BIT dataset.} The BIT dataset \cite{DBLP:conf/eccv/KongJF12} contains 8 types of interactions: bend, box, handshake, hifive, hug, kick, pat and push, all the activities are performed by a pair of actors. Each activity contains 50 video sequences, i.e., totally 400 videos in the dataset. Following~\cite{DBLP:conf/eccv/KongJF12}, a random subset containing 272 videos is used for training, and the remaining 128 videos are used for testing.

\begin{figure*}[t]
\centering
\subfigure[UTI \# 1]{
\label{fig:uti1}
\includegraphics[width=0.315\linewidth]{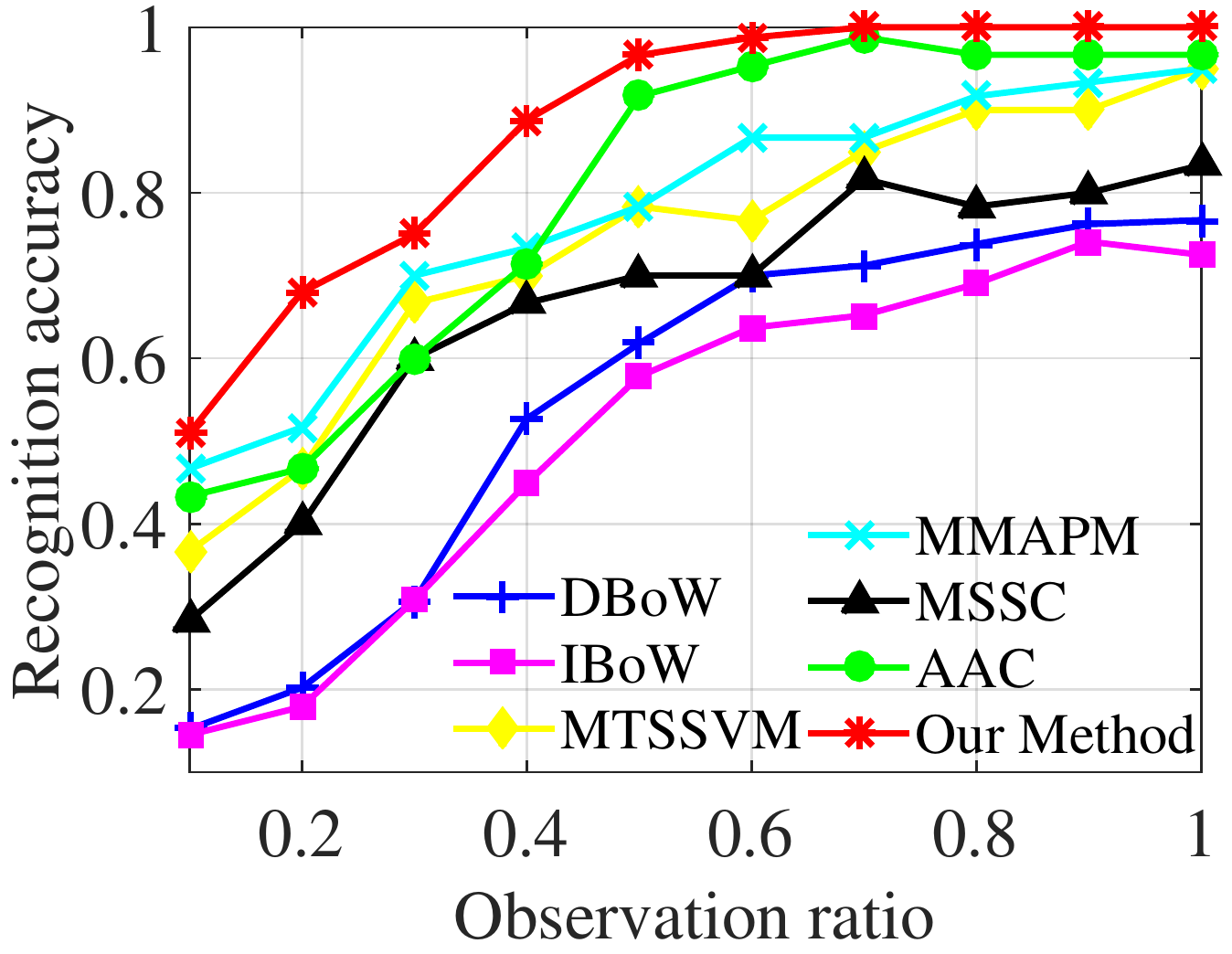}}
\hspace{1mm}
\subfigure[UTI \# 2]{
\label{fig:uti2}
\includegraphics[width=0.315\linewidth]{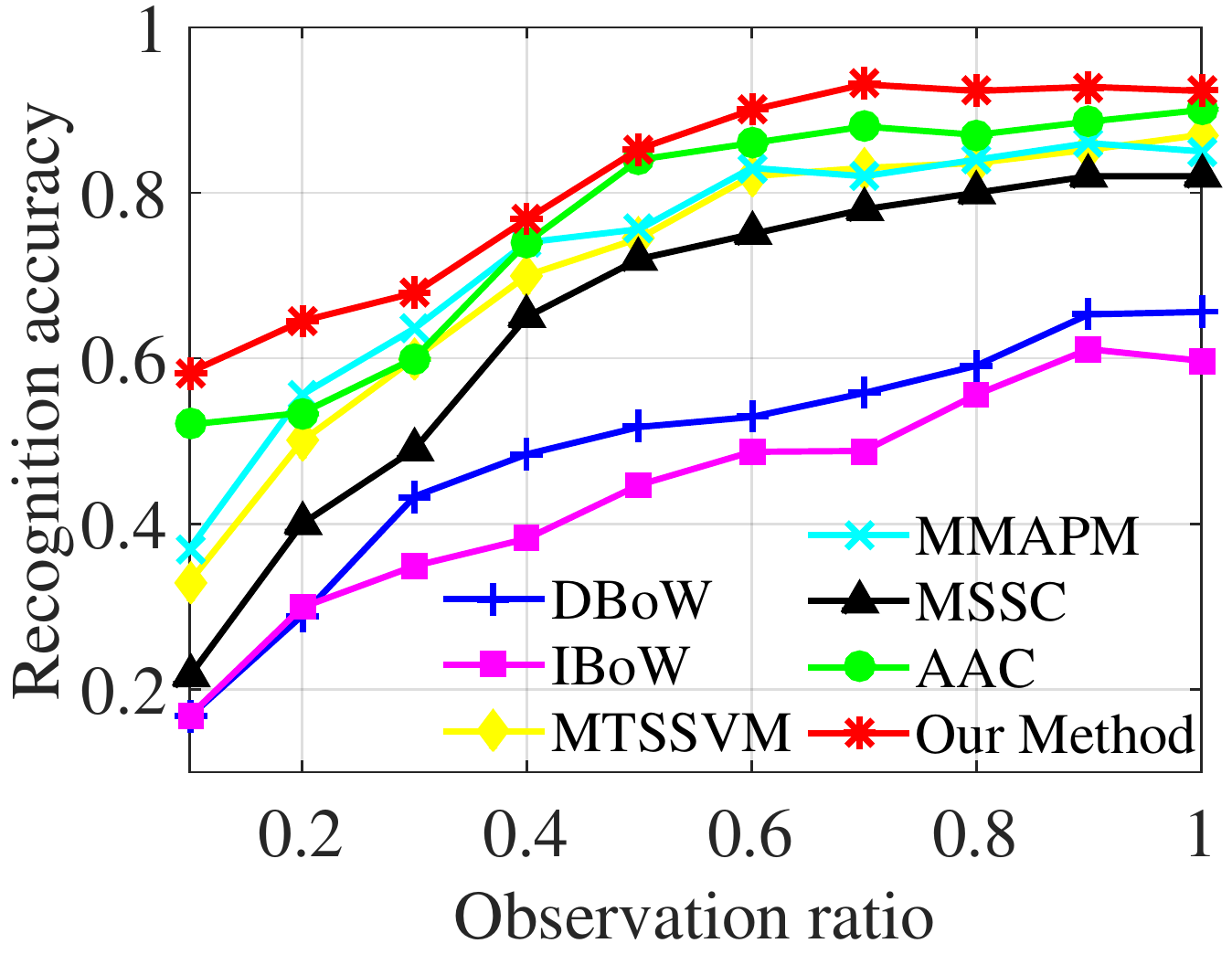}}
\hspace{1mm}
\subfigure[BIT]{
\label{fig:bit}
\includegraphics[width=0.315\linewidth]{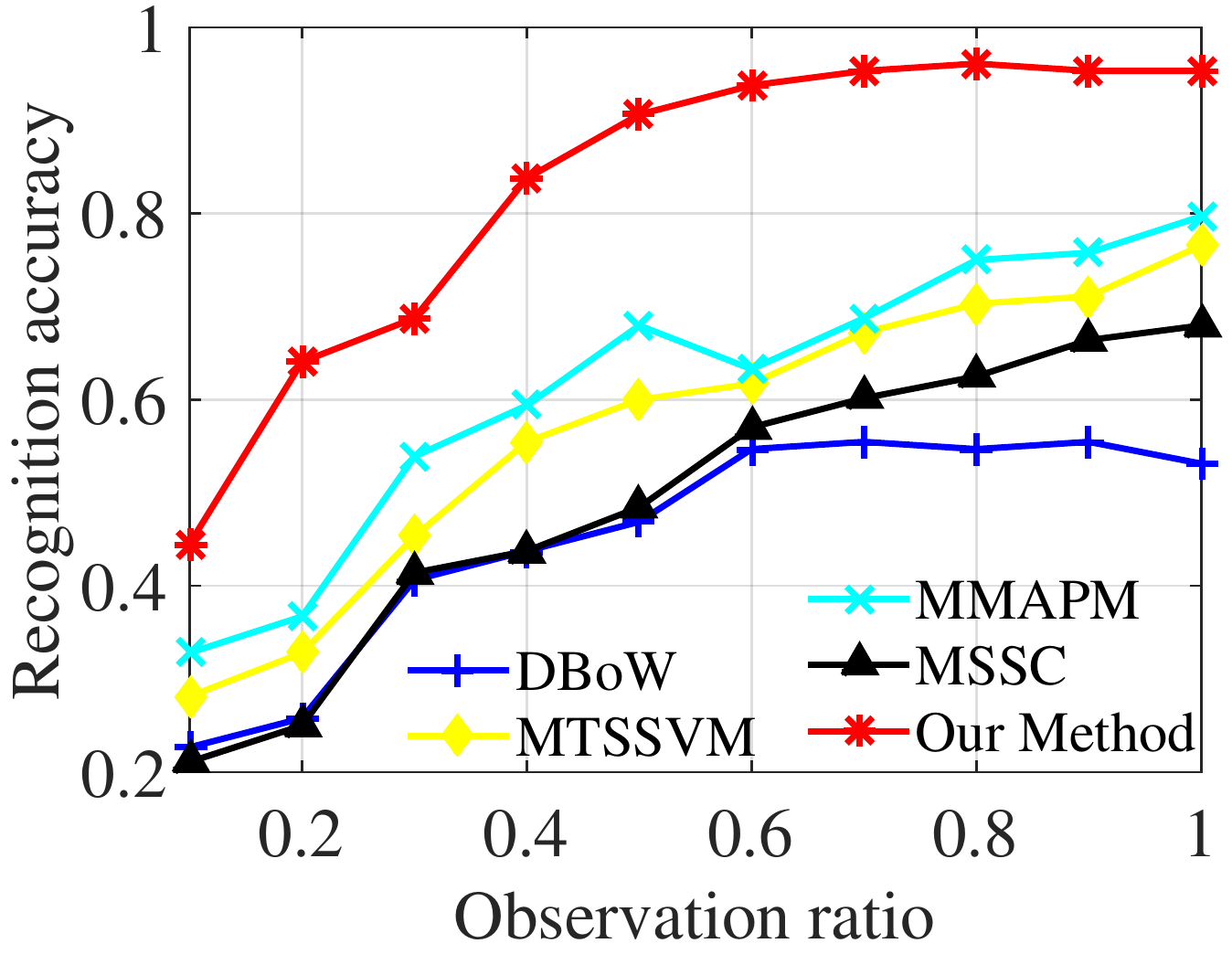}}
\hspace{1mm}
\label{fig:results}
\vspace{-5mm}
\caption{Prediction results on UTI \#1, \#2, and BIT dataset. Our method means the result achieved by our tri-coupled relative attention network on top of optical flows.}
\vspace{-2mm}
\end{figure*}

\subsection{Implementation Details}
The implementation of the proposed networks are based on Caffe \cite{jia2014caffe}. The LSTM layer contains 512 hidden units, and a dropout layer is placed after it to avoid over-fitting. To increase training instances and to make our model applicable for sequences of variable length, we randomly extract subsequences of fixed length $L$ ($L=10$ in our experiments) for training.
To train the LSTM networks, the original learning rate is initialized as $0.001$, and the learning rate is decreased to $\frac {1}{10}$ of the original value after each 10 epochs. The whole training phase includes 30 epochs. The complete duration of training time is about 12 hours on a Titan X GPU.
During testing, we extract the subsequences in the testing video with a stride of 5, and averaging their classification score as prediction.
We test our network on top of both RGB frames and optical flows. The optical flow is computed using the approach of \cite{DBLP:conf/eccv/BroxBPW04}.
As \textit{point} action in the UTI dataset is a single action, we duplicate the image as input for the networks that require both subjects, i.e., the naive fusion network, the coupled network and the tri-coupled network.

\subsection{Results on UTI Dataset}
\begin{table}[]
\centering
\caption{Activity prediction performance on UTI \#1 dataset}
\label{result:uti1}
\begin{tabular}{c|c|c}
\hline
Methods             & OR=0.5          & OR=1           \\ \hline
\textbf{Our method} & \textbf{96.7\%} & \textbf{100\%} \\
DBoW \cite{DBLP:conf/iccv/Ryoo11}               & 70.0\%          & 85.0\%         \\
IBoW  \cite{DBLP:conf/iccv/Ryoo11}              & 65.0\%          & 81.7\%         \\
MTSSVM  \cite{DBLP:conf/eccv/KongKF14}            & 78.33\%         & 95.00\%        \\
MMAPM  \cite{DBLP:journals/pami/KongF16}             & 78.33\%         & 95.00\%        \\
MSSC  \cite{DBLP:conf/cvpr/CaoBBNYMLDSW13}              & 70.0\%          & 83.3\%         \\
AAC    \cite{DBLP:conf/iccv/XuQM15}             & 91.67\%         & 96.67\%        \\ \hline
\end{tabular}
\vspace{-3mm}
\end{table}

\begin{table*}[]
\centering
\caption{Interaction prediction accuracies on UTI \#2 dataset with different observation ratio (OR).}
\label{table:uti2}
\begin{tabular}{c|cccc|cccc}
\hline
\multirow{2}{*}{Methods} & \multicolumn{4}{c|}{RGB inputs}                                        & \multicolumn{4}{c}{Optical flows}                                    \\ \cline{2-9}
                         & OR=0.1          & OR=0.2          & OR=0.5          & OR=1            & OR=0.1          & OR=0.3          & OR=0.5          & OR=1            \\ \hline
Global LSTM              & 16.7\%          & 33.3\%          & 51.7\%          & 63.3\%          & 16.7\%          & 58.3\%          & 70.0\%          & 76.7\%          \\
Naive fusion             & 23.3\%          & 38.3\%          & 58.3\%          & 68.3\%          & 30.0\%          & 61.7\%          & 71.7\%          & 80.0\%          \\
Coupled network        & 16.7\%          & 33.3\%          & 50.0\%          & 58.3\%          & 21.7\%          & 58.7\%          & 68.3\%          & 75.0\%          \\
Tri-coupled network      & 35.0\%          & 48.3\%
    & 65.0\%          & 73.3\%          & 53.3\%          & 65.0\%          & 81.7\%          & 90.0\%            \\
\textbf{Our method}               & \textbf{36.7\%} & \textbf{53.3\%} & \textbf{71.7\%} & \textbf{76.7\%} & \textbf{58.3\%} & \textbf{68.3\%} & \textbf{85.0\%} & \textbf{91.3\%} \\ \hline
\end{tabular}
\vspace{-4mm}
\end{table*}

The proposed tri-coupled relative attention network is compared with some leading approaches on interaction prediction. (1) Bag-of-words based methods: DBow and IBoW \cite{DBLP:conf/iccv/Ryoo11}; (2) Sparse coding based method: MSSC \cite{DBLP:conf/cvpr/CaoBBNYMLDSW13}; (3) Max margin structure SVM based methods: MTSSVM \cite{DBLP:conf/eccv/KongKF14} and MMAPM \cite{DBLP:journals/pami/KongF16}; and (4) discriminative patch based method: AAC \cite{DBLP:conf/iccv/XuQM15}.
The comparative results on UTI \#1 is shown in Figure \ref{fig:uti1}, and the quantitative results with observation ratio 0.5 and 1 are shown in Table \ref{result:uti1}. We report our best performance with tri-coupled relative attention network on top of optical flow inputs. Our method achieves favorable performance compared to other methods. It's remarkable that our tri-coupled structure achieves 100\% recognition accuracy when the observation ratio is larger than 0.6. This is better than the previous state-of-the-art method \cite{DBLP:conf/iccv/XuQM15}, which also achieves remarkable performance on this dataset, i.e., 91.67\% and 96.67\% for half video and full video. We further notice that our results are significantly higher than DBow, IboW \cite{DBLP:conf/iccv/Ryoo11} and other encoding based models. This is mainly because that tri-coupled network explicitly employs the interactive information, while most other methods only rely on the global information.

Comparative results on UTI \#2 are displayed in Figure \ref{fig:uti2}. We notice that other methods have significant lower prediction accuracies compared to the results on UTI \#1, due to more complex backgrounds and more camera jitter. Even the discriminative patch based method AAC \cite{DBLP:conf/iccv/XuQM15} suffers from about 10\% decrease. Compared to these methods, our tri-coupled relative attention model achieves better performance, more than 90\% prediction accuracies when observation ratio is larger than 0.5, which is higher than other methods. This well demonstrates the robustness of the proposed method in existence of noise, and it is mainly due to our relative attention network, which is able to attend to discriminative regions on each interactive subject.

\textbf{Component analysis}. Our framework consists of two major components: the tri-coupled network and the relative attention network. To evaluate the effectiveness of each component, we compare our network with some baseline structures introduced in Section \ref{sec:method}: (1) global LSTM network; (2) naive fusion network; (3) coupled network; and (4) tri-coupled network without relative attention.
The results on UTI \#2 dataset with both RGB inputs and optical flow inputs are shown in Table \ref{table:uti2}, we have three observations.
First, our baseline networks with optical flows achieves much better performance than the baseline methods using RGB frames. This is mainly because that the motion information contained in optical flows is crucial for identifying the actions.
Second, we note that the naive fusion method only achieves marginally increase to the performance compared to global LSTM network, for both optical flows and RGB frames. This is because that some motion patterns of individual subjects can be very similar though different interactions, which may even provide negative information for prediction. e.g., the dodge motion occurs in both kick and punch, it will be difficult to make a prediction when observing such pattern.
Last but not least, the tri-coupled network brings significant performance gain to the above baseline methods, especially in the case of high observation ratios.
When embedded with the relative attention network, the performance is further improved.
This demonstrates the effectiveness of the proposed tri-coupled network as well as the relative attention network.

\subsection{Result on BIT Dataset}
The results on BIT dataset are shown in Figure \ref{fig:bit}. All the other methods get worse results compared to the results on UTI datasets, due to the fact that BIT dataset contains more category of interactions, and the videos in this dataset are with more complex backgrounds and sometimes with heavy occlusion. Therefore, the compared methods \cite{DBLP:conf/iccv/Ryoo11,DBLP:journals/pami/KongF16,DBLP:conf/cvpr/CaoBBNYMLDSW13,DBLP:conf/eccv/KongKF14} achieve less than 80\% prediction accuracies even with full observation, which is far away from real-world applications.
While our network achieves more than 90\% accuracy when only only half the video is observed, which outperforms the compared methods by a large margin (more than 10\% with any observation ratio). This is because of the effectiveness of the relative attention module, which is able to attend to the discriminative regions in the scene, thus make the proposed method more robust to occlusions.

\begin{figure}
\centering
\subfigure[Bow]{
\label{fig:map1}
\includegraphics[height=1in]{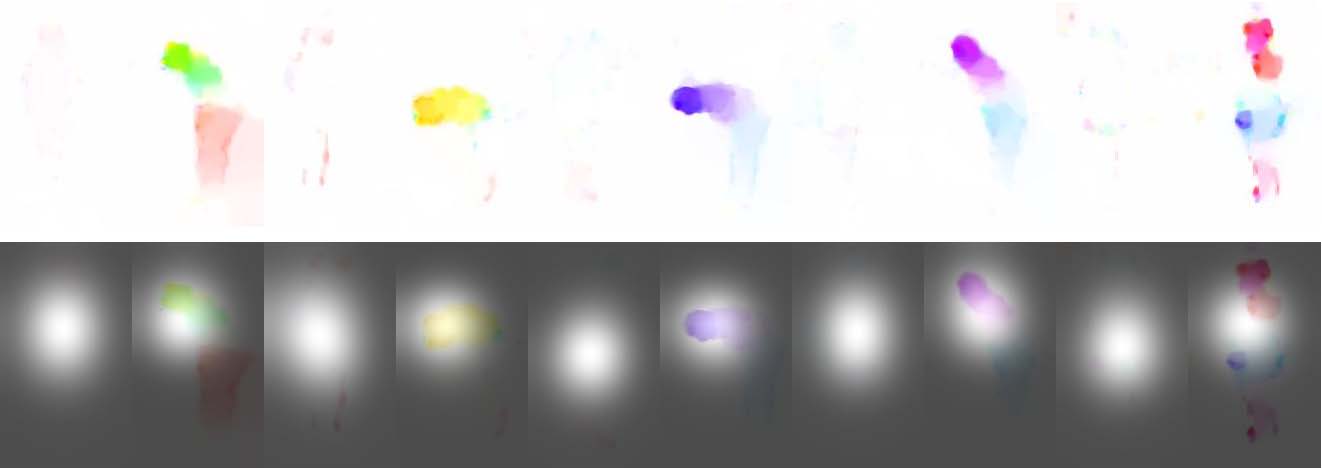}}
\vspace{-3mm}
\subfigure[Handshake]{
\label{fig:map2}
\includegraphics[height=1in]{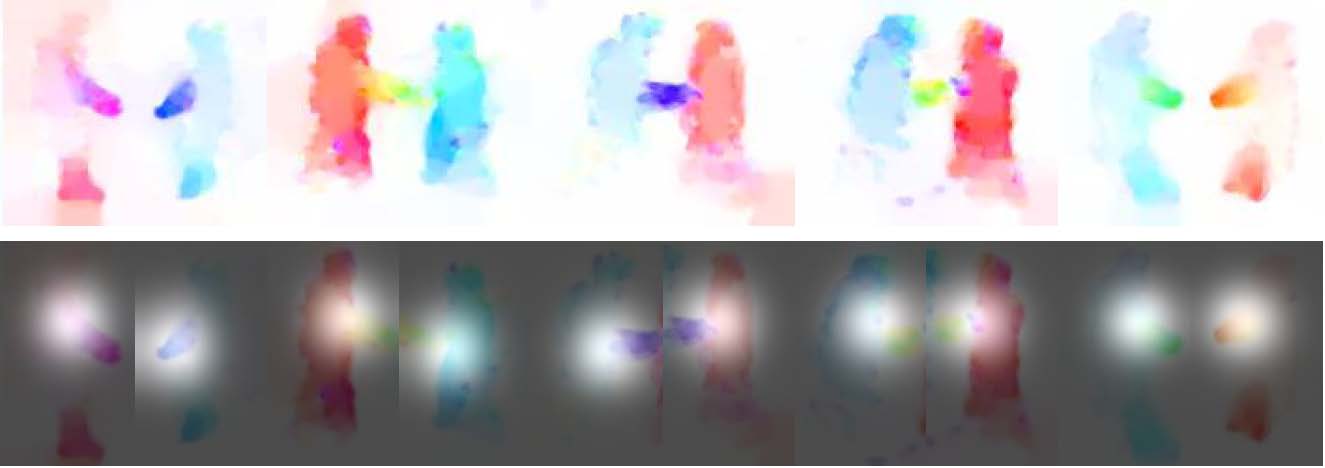}}
\vspace{-1mm}
\subfigure[Kick]{
\label{fig:map3}
\includegraphics[height=1in]{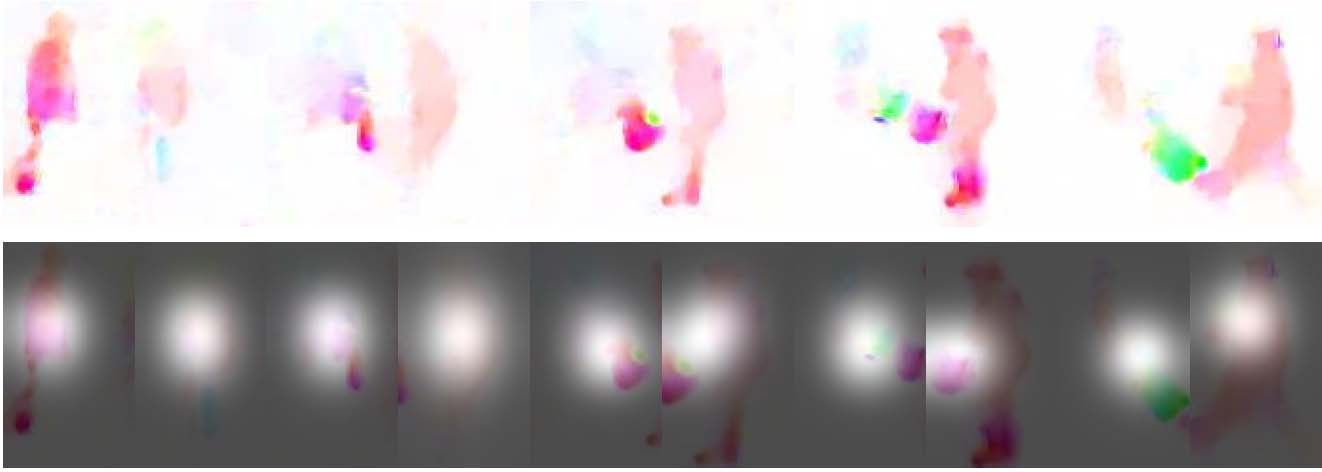}}
\vspace{-4mm}
\caption{
Examples of attended regions on optical flows.
}
\label{fig:qualitative_results}
\vspace{-3mm}
\end{figure}

\subsection{Qualitative Results}
Figure \ref{fig:qualitative_results} visualizes the attended regions generated by our relative attention model.
As the best performance of our method is achieved upon optical flows, the visualization is based on optical flow images.
The first example illustrates the interaction of \textit{bend}. It's easy to notice that the major subject is on the right side, and the subject on the left nearly has no movement during the interaction. For the subject on the right, we can find very strong correlation between the attended regions and the movements of the upper part of the body.
The second example depicts two people shaking their hands. Both subjects are involved during the interaction, and they share similar behaviours: stepping forward and reaching out their hands. Our model consistently attends to the arms of both subjects, which shares similar intuition of human cognition.
In the last example, the subject on the left is kicking the right subject. The attended regions are focused on the extended leg of the left subject, and the upper body of the right subject is attended to as he/she falls down.

\section{Conclusion}
In this paper, we propose a tri-coupled relative attention network for human interaction prediction.
Experimental results convincingly demonstrate that the proposed relative attention network successfully predicts informative regions between interacting subjects, which in turn yields superior human interaction prediction accuracy.
Although this paper is explicitly designed to model two subject interaction, our method is easily extendable to model group people interaction. Here is a brief illustration for this generalization. For each subject, the relative attention could be calculated with his/her nearest neighbors. The computational complexity only increases linearly w.r.t. the number of neighboring subjects. Finally, we can aggregate all groups via a LSTM structure to achieve global group activity label prediction.

\section*{Acknowledgments}

The work was supported by State Key Research and Development Program (2016YFB1001003). This work was partly supported by National Natural Science Foundation of China (NSFC61502301, NSFC61521062), China's Thousand Youth Talents Plan, the 111 Project (B07022) and the Shanghai Key Laboratory of Digital Media Processing and Transmissions.
\bibliographystyle{named}
\bibliography{ijcai17}

\end{document}